\documentclass[10pt,twocolumn,letterpaper]{article}

\usepackage{cvpr}
\usepackage{times}
\usepackage{epsfig}
\usepackage{graphicx}
\usepackage{amsmath}
\usepackage{physics}
\usepackage{amssymb}
\usepackage{url}
\usepackage[utf8]{inputenc}
\usepackage{multirow}

\usepackage[pagebackref=true,breaklinks=true,letterpaper=true,colorlinks,bookmarks=false]{hyperref}

\cvprfinalcopy % *** Uncomment this line for the final submission

\def\cvprPaperID{****} % *** Enter the CVPR Paper ID here

\ifcvprfinal\fi
\begin{document}

%%%%%%%%% TITLE
\title{Ego-Downward and Ambient Video based Person Location Association}

\author{Liang Yang $^{1}$, Hao Jiang $^{2}$, Jizhong Xiao $^{1}$, Zhouyuan Huo$^{3}$\\
$^{1}$ Robotics Lab, The City College of New York, City University, New York, NY\\
$^{2}$ Microsoft, Redmond, USA\\
$^{3}$ University of Pittsburgh, Pittsburgh, USA\\
{\tt\small lyang1,jxiao@ccny.cuny.edu, jiang.hao@microsoft.com,zhouyuan.huo@pitt.edu}
}

\maketitle
%\thispagestyle{empty}

%%%%%%%%% ABSTRACT
\begin{abstract}

Using an ego-centric camera to do localization and tracking is highly needed for urban navigation and indoor assistive system when GPS is not available or not accurate enough. The traditional hand-designed feature tracking and estimation approach would fail without visible features. Recently, there are several works exploring to use context features to do localization. However, all of these suffer severe accuracy loss if given no visual context information. To provide a possible solution to this problem, this paper proposes a camera system with both ego-downward and third-static view to perform localization and tracking in a learning approach. Besides, we also proposed a novel action and motion verification model for cross-view verification and localization. We performed comparative experiments based on our collected dataset which considers the same dressing, gender, and background diversity. Results indicate that the proposed model can achieve $18.32 \%$ improvement in accuracy performance. Eventually, we tested the model on multi-people scenarios and obtained an average $67.767 \%$ accuracy.

\end{abstract}

%%%%%%%%% BODY TEXT
%-------------------------------------------------------------------------

\section{Introduction}

%ToDo:

In recent years, accurate localization and consistent tracking in a large crowd, including the shopping mall, urban street, airport, and public park, possibly involved with interaction for identification of specific requests, are extensively needed, especially for visually impaired people \cite{xiao2015assistive} and urban navigation with high accuracy localization request \cite{googleArNavigation}. However, the requirement of large storage for pre-recorded feature map \cite{klingensmith2015chisel} limits its usage in a large open area. Besides, the problem of view block and the lack of static features for tracking also make it harder to be implemented in urban areas \cite{bresson2017simultaneous}. It is highly required to have a stable and mobile capable approach to solve this problem in a high accuracy.

\begin{figure}[!th]
\begin{center}
   \includegraphics[width=1.0 \linewidth]{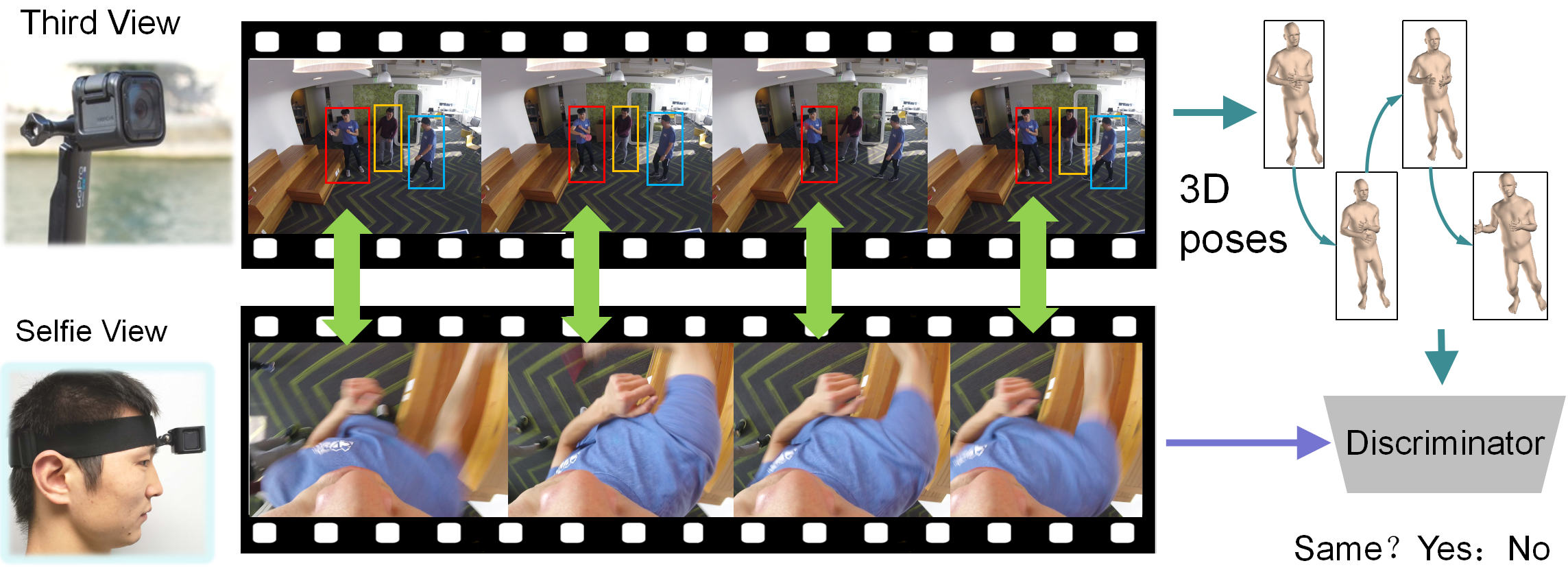}
\end{center}
   \caption{ The architecture of the proposed ego-downward and third view assist system. The ego-downward camera (not able to be blocked) is used to localize the person in the third view.}
\label{fig:cameraSystemIntroduction}
\end{figure}

In this paper, we propose to use a mobile camera and a static third view camera system as illustrated in Fig.\ref{fig:cameraSystemIntroduction} to address this problem. We assume that the person wears a head-mount camera which observing a downward narrow area (a case for VR game headset). We aim to verify how an ego downward camera and a third view camera can be used for verification and localization in the wild.

Note that there are some existing works on third and ego-centric view matching analysis for the human. All of these approaches, however, focusing on using two streams siamese or triplet network structure \cite{fan2017identifying, sigurdsson2018actor, ardeshir2018integrating} to learn to identify between third and ego view. In these models, the most recent approaches including 3D convolutional neural network \cite{tran2015learning,wang2016temporal,qiu2017learning} and segmental consensus for cross-domain verification \cite{wang2016temporal, fan2017identifying, sigurdsson2018actor} are deployed. However, these approaches cannot generalize the knowledge of pose and motion for human tracking and cross view verification. Thus, pure visual features are not capable to model the variance of the human action across views toward tracking, especially the ego-downward view can only visualize the human itself.

Unlike the top-and-forward view \cite{ardeshir2018integrating} and third-forward view \cite{fan2017identifying} cases, the ego-downward mounting faces the following challenges : 1) appearance verification across different views does not hold under this situation since it is not pointing out to scenario; 2) clothes texture verification will not work since in large crowd there should have the similar dressing or occlusion; 3) the same action with different initial pose state (in world coordinate system) will also mislead the model since the ego-downward frames will not tell the difference (in Fig.\ref{fig:pose_variance_as_info}). Thus, using a general siamese or triplet model to correlate the two views with temporal and spatial information would fail \cite{fan2017identifying}. Moreover, the graph solution using relative view insight will not happen under this situation \cite{ardeshir2018integrating}.

In this paper, we proposed a novel action and motion feature based model to address these challenges. Our key learning is that ego view can always visualize part of the body, and thus can help to estimate the pose variance \cite{xu2018mo2cap2} and body motion. Our main contribution is to learn action and 3D motion feature for cross view verification, via taking advantage of the third view person tracker and 3D pose estimation. It can be summarized as follows:
\begin{itemize}
\item Firstly, we introduce to use a Yolo \cite{redmon2018yolov3, Bewley2016_sort} based tracker to do human tracking to provide enough continuous sequence. Meanwhile, we perform 3D pose estimation for the ego and the third view alignment.

\item Secondly, we propose a novel action and motion verification and tracking model for cross views in Section.\ref{Sec:method}. Using this model, the ego-view pose and transformation can be aligned into the third view, which is more sufficient for regression.

\item Finally, we build a comprehensive dataset and validate our proposed method in Section.\ref{Sec:experiments}. Experimental results demonstrate that the proposed method achieves higher accuracy and is robust to body context as well as the background.

\end{itemize}

%-------------------------------------------------------------------------

\section{Related Works}

\textbf{Ego-centric and Third View Joint Modeling} - The problem of associating first (mobile) and third (static) view was firstly discussed in \cite{alahi2008object} to improve the object detection accuracy in the third view. Authors in \cite{soran2014action} discussed the problem of using the egocentric and third view camera to perform action recognition, which addressed the fact that egocentric cameras benefit the recognition. In \cite{ardeshir2016ego2top}, the authors correlated the first view and third view firstly. The authors proposed a 'Graph' representation for temporal and spatial matching. In \cite{fan2017identifying}, the authors solved the task to localize the person in the third view if given the both the third and ego camera frames. In this paper, spatial-domain semi-siamese, motion-domain semi-siamese, dual-domain semi-siamese, and dual-domain semi-triplet networks are well studied. Besides correlation method discussion, authors in \cite{sigurdsson2018actor} released "Charades-Ego Dataset" to study the problem of daily human activity study and provide the baseline of performing basic frame-to-frame association. These works mainly consider context features as the main clue, and they did not consider pose features and motion (odometry) feature for verification. Besides, our work differs from other work that we perform an association of downward view and third static view, which could help to increase the robustness of tracking.

\begin{figure*}[!th]
\begin{center}
   \includegraphics[width=1.0 \linewidth]{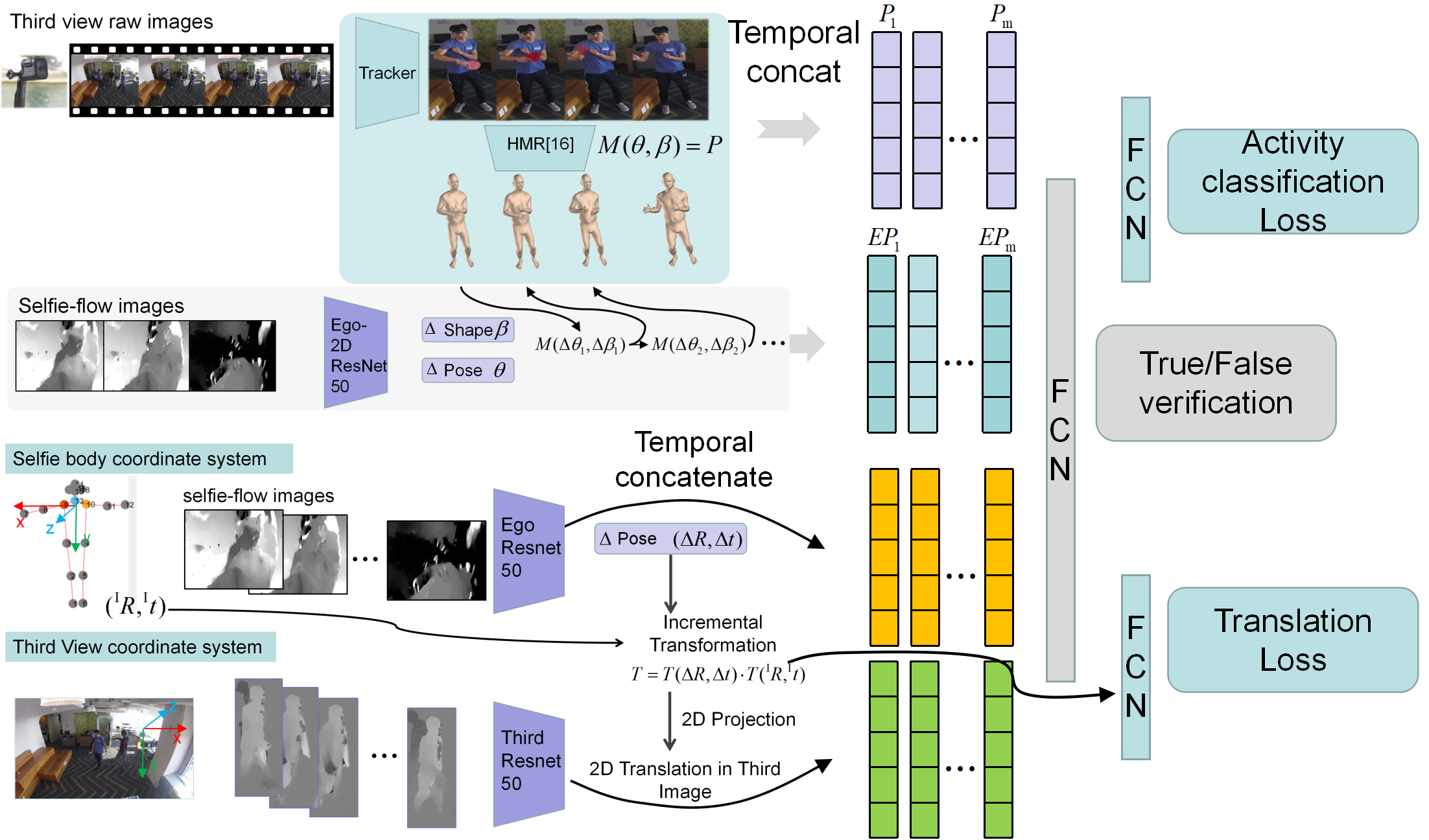}
\end{center}
   \caption{The Ego-downward and third view verification and tracking (ETVVT) model proposed in this paper. The model is learning in a joint approach of ego-downward view and third view using motion and action feature.}
\label{fig:networkarchitecture}
\end{figure*}

\textbf{Temporal and Spatial Model for action Learning} - Temporal information was first introduced to solve action recognition in \cite{tran2015learning}, where a $3D$ convolutional operation with $3D$ max-pooling were first discussed which greatly improved the performance of learning temporal features. Then, a ResNet \cite{he2016deep} based 3D convolutional neural network is proposed in \cite{qiu2017learning}  to achieve higher accuracy using a smaller model. Spatial information is commonly used in detection and correlation \cite{johnson2015image} using context information or objects information. For egocentric and third view matching the task, temporal and spatial information is first discussed in \cite{ardeshir2016ego2top} using a naive concatenation approach. Then, work \cite{fan2017identifying} proposed using $3D$ convolutional approach to perform the temporal learning. However, none of the above method learn the pose information in temporal or spatial domain to perform association. Current success in human pose detection \cite{cao2017realtime} enables the learning of action in a graph convolution manner \cite{yan2018spatial} in both temporal and spatial domain.

\textbf{Learning for Localization} - RGB-D images based localization \cite{shotton2013scene} is the first localization approach used widely. Then, the first learning approach toward end-to-end localization is proposed in \cite{kendall2015posenet}. In order to address the sequence continuous constraints, authors in \cite{clark2017vidloc} proposed recurrent network to enable smooth localization. \cite{valada2018deep} demonstrated how to multitask which incorporate visual odometry prediction and global localization can relieve requiring of a huge dataset and achieve higher localization accuracy as well. Lately, authors in \cite{mapnet2018} introduced almost the same idea as \cite{valada2018deep} of performing multitask toward localization, while this work differs in introducing both pose loss and velocity loss to increase the convergence of the model. Tracking is a traditional topic in both computer vision and robotic area \cite{watada2010human}, and later learning approach has been successfully demonstrated with real-time performance \cite{held2016learning}.

%-------------------------------------------------------------------------

\section{Method}
\label{Sec:method}

The proposed model is illustrated in Fig.\ref{fig:networkarchitecture}. It contains two sub-blocks, which are action sub-model and motion sub-model. For action sub-model, given an third image of a person at time $k$, we performed 3D pose estimation to obtain $^{T}\c{P}_{k}$ to initialize ego-downward view frame at time $k$. Then, at time $k+1$, the third view still performs 3D pose estimation $^{T}\c{P}_{k+1}$, while ego-downward view tells pose variation $\Delta ^{E}\c{P}_{k+1}$. Thus, we can obtain two pose sequences as $\{^{T}\c{P}_{k}, ^{T}\c{P}_{k} +  \Delta ^{E}\c{P}_{k+1} \}$ for ego-downward view, and $\{ ^{T}\c{P}_{k}, ^{T}\c{P}_{k+1}\}$ for third view. The two pose sequences should be the same. For motion sub-model, 3D joints of human body can provide the transformation, $T_{init}=(^{T}R, ^{T}t)$, between Ego and third view. Then, at time $k+1$, the ego model model predicts the transformation $\Delta T = (\Delta R, \Delta t)$ from $k$ to $k+1$. In $SE(3)$, the transformation of $k+1$ is, $T_{k+1}  = T_{init} \cdot \Delta T$, and we can have $T^{Ego} = \{T_{init},  T_{k+1}, ..., T_{l}\}$ for all $l$ consecutive frames. Mean while, the third view directly predicts the relative translation in image domain as, $(\Delta x, \Delta y)$. Then, the third view translation is, $^{trd}t=[0, 0; \Delta x, \Delta y; ...;\sum_{i=1}^{7}\Delta x, \sum_{i=1}^{7}\Delta y]$. The two translation should also be the same in third view. It should be noted that the sequence translation is represented in third view coordinate system which is the default world frame in this paper (as illustrated in Fig.\ref{fig:networkarchitecture}).

\subsection{Learning Action Feature by Applying 3D Pose}
\label{sec:Activity_section}

\begin{figure}[!th]
\begin{center}
   \includegraphics[width=1.0 \linewidth]{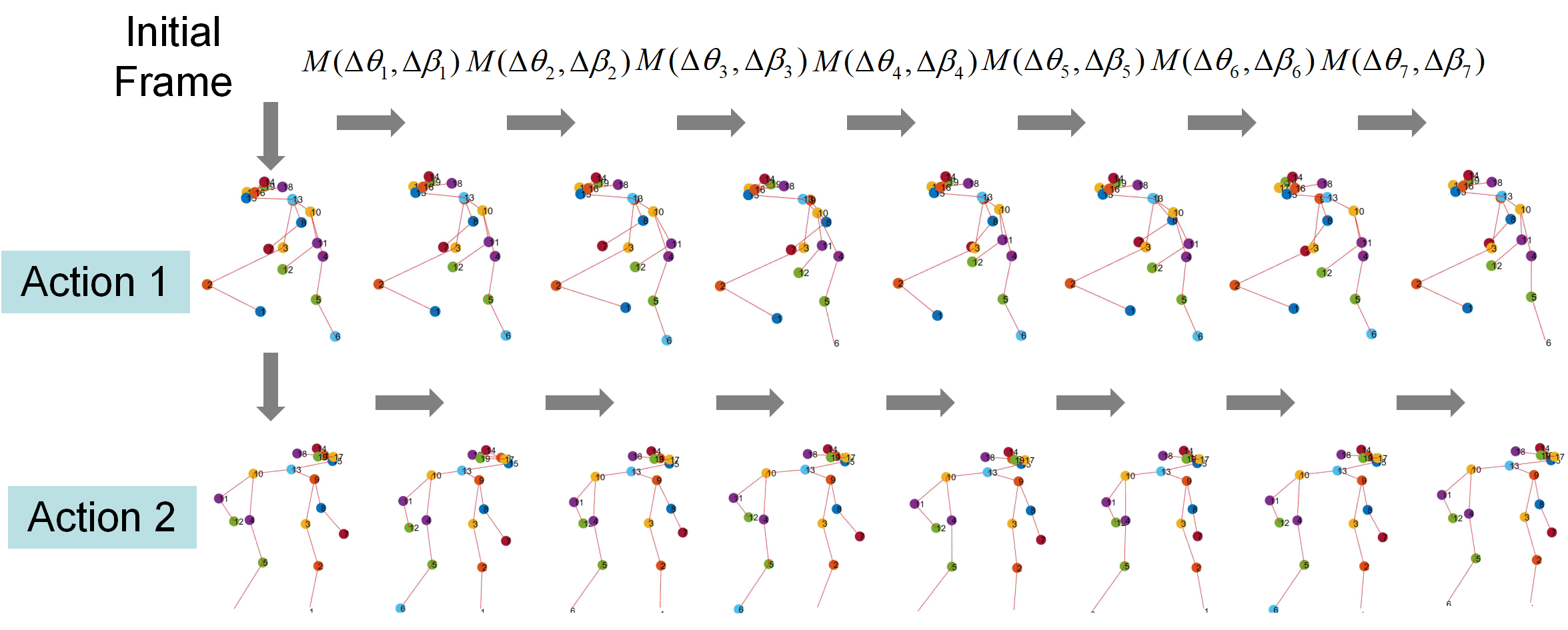}
\end{center}
   \caption{Given the same $7$ continuous 3D pose variations and two different initial 3D poses. We can find the resulting action are totally different and highly related to the initial 3D pose. For ego-downward frames, they can only be used to predict the pose variation, and thus the view verification must consider the initial 3D pose. }
\label{fig:pose_variance_as_info}
\end{figure}

\textbf{Preliminary Definitions}: $ \ $ We represent the human pose using 3D joints as Skinned Multi-Person Linear (SMPL) \cite{SMPL2015} model and unlike the original 24 joints, we use the 19 joints which are defined in \cite{kanazawa2018end} as: $1:Right  \ ankle, \ 2: Right \  knee, \ 3:Right \  hip, \ 4:Left  \ hip,\ 5:Left \  knee,\ 6:Left  \ ankle,\ 7: Right  \ wrist,\ 8:Right \  elbow,\ 9:Right \  shoulder,\ 10:Left \  shoulder,\ 11:Left  \ elbow,\ 12:Left \  wrist,\ 13:Neck,\ 14:Head  \ top,\ 15:nose,\ 16:left  \ eye,\ 17:right  \ eye,\ 18:left \  ear,\ 19: right  \ ear$. For the SMPL model, it factors the human body into shape $\beta $ - how individuals vary in height, weight, body proportion and poses $\theta$ - the 3D surface deforms with articulation. The whole model consists of $N=6890$ vertices to form a 3D mesh which is continuous quad structure, and represented as $M(\beta, \theta; \Phi): R^{|\theta| \times |\beta|} \mapsto R^{3N}$.

The tracked person in the third view with a bounding box is cropped out in original RGB-image as $^{cr}Y$ and the optical flow images as $^{cr}Y^{fl}$. The cropped third view images are directly used to estimate the 3D pose $\c{p}$ with $19$ joints. In this paper, we use $8$ consecutive pose to represent an action.

\textbf{Learning Third View Action} $ \ $ We first classify the 3D poses $\c{p}$ over $80000$ poses into $400$ clusters as $L$. For a consecutive $8$ frames, $^{cr}Y = \{^{cr}I_{i}|i=0,1,...,7\}$ and its corresponding 3D action cluster label $^{trd}L \mapsto K-means(\{ ^{trd}\c{p}_{i}| i=0,1,...,7\})$. Each third view clip has a dimension of $8 \times W \times H \times C$, with $C$ Channels, $W$ width, $H$ height, and $8$ frames. The third view poses network architecture is composed of a 3D ResNet-18, with a total 4 blocks. The first three blocks are with a max-pooling of $2 \times 2 \times 2$ in both spatial and temporal channels, and there is no temporal pooling with the four blocks. We only perform a 2D convolution for feature extraction. 3D ResNet doubles the depth while the dimension decreased starting from $64$ for the first block and $512$ for the fourth block. The final output after average pooling is a $512$ dimensional vector. 3D ResNet-18 then connects with a fully-connected network with a total $3$ layers to perform action prediction.

\textbf{Ego-downward View Pose Variation Prediction Model}  $ \ $ One learning is illustrated in Fig.\ref{fig:pose_variance_as_info}. Given two initial frames with poses $\c{p}^{1}$ and $\c{p}^{2}$. Also, the consecutive $8$ frames pose variation is given as $\Delta \c{P} = \{\Delta \c{p}_{i}|i=1,2,...,7\}$. Then, we can obtain the corresponding 3D action sequence as $\c{A}^{1} = \{ \c{p}^{1}, \c{p}^{1} + \Delta \c{p}_{i}, ...,  \c{p}^{1} + \sum_{i=1}^{7} \Delta \c{p}_{i} \}$ and $\c{A}^{2} = \{ \c{p}^{2}, \c{p}^{2} + \Delta \c{p}_{i}, ...,  \c{p}^{2} + \sum_{i=1}^{7} \Delta \c{p}_{i} \}$. It can clear conclude from Fig.\ref{fig:pose_variance_as_info} that the two action $\c{A}^{1}$ and $\c{A}^{2}$ are different actions in global view (third view), even given the same ego view action.

For a clip of ego-downward flow images $X = \{ ^{ego}I^{flow}_{i}|i=1,..7 \}$, which can obtain major part of the body motion (It is illustrated in Fig.\ref{fig:networkarchitecture}). The configuration of the selfie model is a 2D ResNet-50. The input is $W \times H \times C$ image with channel $C = 2$, width and height $W=H=112$ as the original model. The output of the ResNet is $2048$ dimensional vector. Then, we introduce to directly use an iterative fully connection network to estimate the shape and pose with $\Delta\beta = \Delta\beta + \Delta \Delta\beta$ and $\Delta\theta = \Delta\theta + \Delta \Delta \theta$, where $\Delta \Delta$ is the variation of the iterative error.

Thus, ego-downward pose variation model directly estimate the pose error between two consecutive frames, $\Delta P_{smpl} = (\Delta\beta, \Delta \theta)$. Given the initial 3D pose as $(\beta, \theta)$, we can thus have the 3D joint pose for a selfie clip as $^{ego}\c{P} = \{ M(\beta, \theta), M(\beta + \Delta\beta_{1}, \theta + \Delta \theta_{1}), ..., M(\beta + \sum_{i=1}^{7}\Delta\beta_{i}, \theta + \sum_{i=1}^{7}\Delta \theta_{i})\}$.

\subsection{Learning Motion for Correlation}

\begin{figure}[!th]
\begin{center}
   \includegraphics[width=1.0 \linewidth]{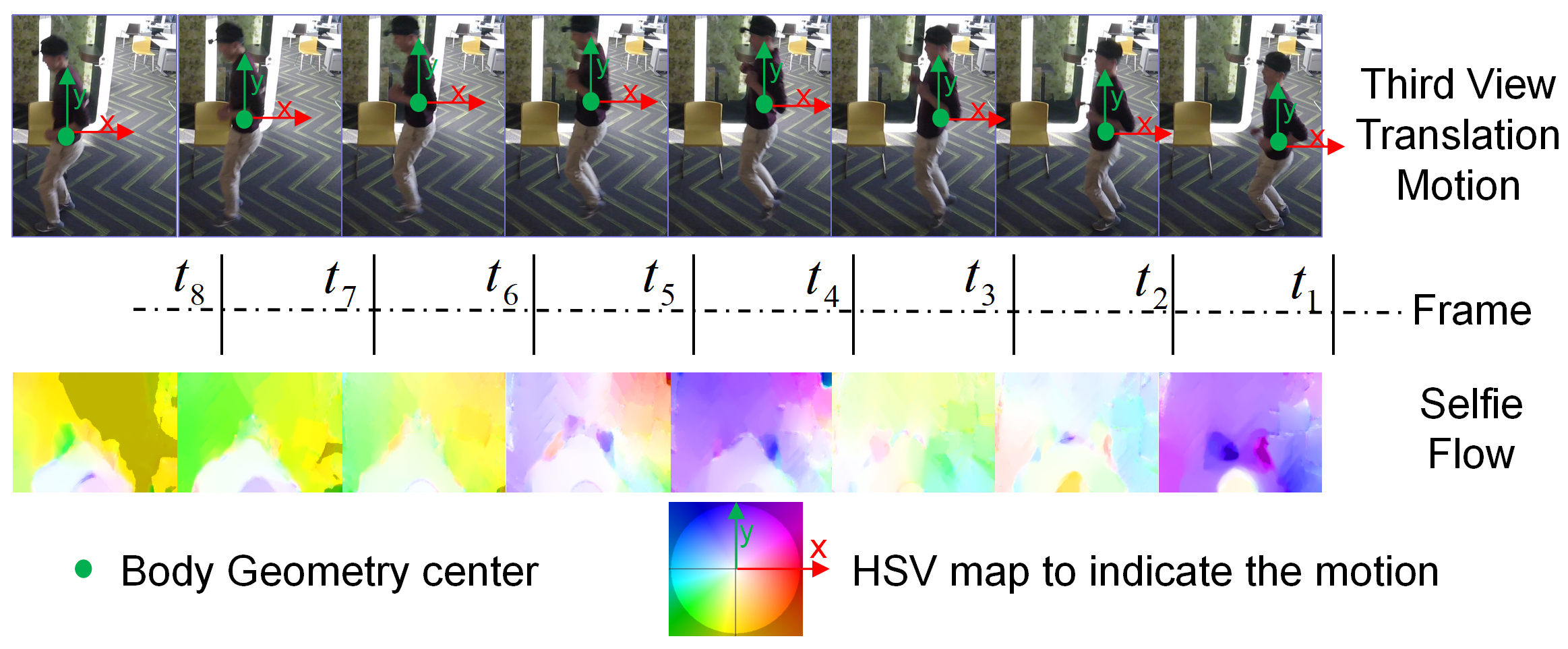}
\end{center}
   \caption{In third view, the motion directly reflect the 2D translation of the body. Meanwhile, the flow image of ego-downward frame tells the motion of the body. }
\label{fig:flow_translation_as_info}
\end{figure}

\textbf{Preliminaries}  $ \ $  In this paper, we also introduce information information, that is, translation to leverage geometric consistency in both third and ego-downward view. It is illustrated in Fig.\ref{fig:flow_translation_as_info}, the third view tracker can generate $m$ bounding boxes for a person $\c{B} = \{(lx_{i}, ly_{i}), (rx_{i}, ry_{i}) | i = 0,1,...,m\}$, then the center (solid green dot) as translation of the sequence in third view image can be described as $\c{B}^{trj} = \{\c{B} - \c{B}_{0} \}$. We can tell that center directly reflects the motion of the person.

\textbf{Learning Third View Translation} $ \ $ To learn third view motion to obtain translation, we introduce 2D $ResNet-50$ and followed by two fully connected layers architecture to predict the frame-to-frame translation. The input is the $7$ consecutive third view cropped flow images $^{cr}Y = \{^{trd}I^{flow}_{i}|i=1,...,7\}$, and the expectation is the tracked bounding box centers sequence $\c{B}^{trj}$. The reason for choosing the flow as input is that the flow image denotes the pixel motion between two frames as, $\dv{I}{x} V_{x} + \dv{I}{y} V_{y} = - \dv{I}{t} \Delta t$. Where $V_{x}$ and $V_{y}$ are the components of velocity in image frame $x$ and $y$ axis of optical flow, $\dv{I}{x}$, $\dv{I}{y}$, and $\dv{I}{t}$ are the derivatives of each pixel in $x, y, t$ direction. It can directly reflect the motion information for prediction.

For each flow frame, the motion model predicts the translation of human in third view image as $(\Delta x, \Delta y)$. In a consecutive $7$ frames of the flow images, the model outputs the frame-to-frame translation as $V=\{ (\Delta x_{i}, \Delta y_{i})| i=1,2...,7 \}$. Thus, the predicted translation in $8$ frames RGB images is, $ [0, 0; \Delta x_{1}, \Delta y_{1}; ...; \sum_{i=1}^{7} \Delta x_{i}, \sum_{i=1}^{7}\Delta y_{i}]$.

\textbf{Learning ego-downward View Translation} $ \ $  Ego-downward motion is highly related to initial pose in the third view, that is, the same motion (transformations with time in third view coordinate system) with different initialization would be total different (in Section.\ref{sec:Activity_section}). The ego-downward view coordinate system is represented by joints $9: Right shoulder, 10: Left shoulder, 13: Neck$ as illustrated in Fig.\ref{fig:networkarchitecture} (the ego-downward body coordinate system block), where $x$ points from left shoulder to right shoulder, $z$ points out and perpendicular to the chest, and $y$ points downward which is perpendicular to $x$ and $z$ axis. In this paper, we deploy $SE(3)$ to represent the transformation $T$ between frames which is consists of a translation $t \in R^{3}$ and a rotation $R \in R^{3 \times 3}$ in 3D space.

Given 3D human body pose $\c{p} = \{(x_{i}, y_{i}, z_{i})| i=1, 2, ..., 19\}$, the center is $\c{p}_{center} = (\c{p}_{9} + \c{p}_{10} +\c{p}_{13})/3$ and the orientation of the person in third view coordinate system is $\vec{r_{p}} = U(cross(\c{p}_{9} - \c{p}_{10}, \c{p}_{13} - \c{p}_{10}))$, where $u$ denotes up direction of the cross product. The transformation between ego and third view then is represented as $^{third}T_{init} = (\vec{r_{p}}, \c{p}_{center})$. For ego motion model, it predicts the frame-to-frame transformation as $T_{t_{k}}^{t_{k+1}} = (\Delta R, \Delta t; 0_{1 \times 3}, 1)$ with ego flow image input, where $\Delta R$ denotes the relative rotation between two ego-downward frames and $\Delta t$ denotes the translation.

In this paper, we use quaternion $q$ to represent the rotation predict as \cite{kendall2015posenet}. However, the rotation difference between any two frame is small enough to represent in error quaternion form \cite{armesto2007fast}, that is, $q_{t_{k+1}} = q_{t_{k+1}|t_{k}} \otimes q_{t_{k}}$. Where, $q_{t_{k+1}|t_{k}}$ is called the error quaternion as:

\begin{equation}
\begin{array}{ll}
  q_{t_{k+1}|t_{k}} &  = exp(\frac{\Delta \theta}{2})  \\
    &  =\begin{cases}  \left[
  \begin{array}{c}
  cos(||\frac{\Delta \theta}{2}||)   \\
  sin(||\frac{\Delta \theta}{2}||) \frac{\Delta \theta}{||\Delta \theta||}
  \end{array}
\right]  &  ||\Delta \theta|| \neq 0 \\
 [1 \ 0 \ 0 \ 0] & ||\Delta \theta|| = 0

   \end{cases}
\end{array}
\end{equation}

Thus, in this paper the ego-downward motion model predicts the quaternion error $\Delta q \in R^{3}$ (which is only 3 parameters) and relative translation $\Delta t \in R^{3}$ with a total $6$ parameters.

\subsection{Training and Regression Details}

Our ETVVT model is composed of action block and motion block. For action block, a siamese structure is introduced of using third view clip and ego-downward to perform action prediction. The Siamese network is also used for learning the motion information for cross view matching. Each block is trained independently and then acts as pre-trained model for ETVVT model.

\textbf{Ego-downward View Action Regression} $ \ $ To learn the action classification in ego-downward view, the input is 3D pose $\c{P}_{init} = M(\beta, \theta)$ and the ego-downward flow clip $X = \{ ^{ego} I_{i}^{flow} |i=1,..7 \}$. The ego-downward action model is supervised to predict the action label using cross entropy loss,

\begin{equation}
    L(X^{a}) = \sum_{i=0}^{399} y_{o,i} log(P_{o,i}),
\end{equation}

$y_{o,i}$ is the binary indicator if the class label $i$ is the correct prediction of current observation and $P_{o,i}$ denotes the corresponding probability.

\textbf{Third View Action Regression} $ \ $ For third view action mode, it directly uses the cropped person sequence $^{cr}Y = \{^{cr}I_{i}|i=0,1,...,7\}$ as input, where $^{cr}Y = (W = 122) \times  (H=122) \times (C=3)$. Then, the fully connected layers predict action label using the 3D $ResNet-18$ features with cross entropy loss,

\begin{equation}
    L(Y^{a}) = \sum_{i=0}^{399} y_{o,i} log(P_{o,i}),
\end{equation}

\textbf{Ego-downward View Transformation Regression} $ \ $  Ego-downward motion model predicts $error \ quaternion \ \Delta q$ and $relative \ translation \ \Delta t$. It can be represented as a transformation, $T_{k}^{k+1} = [r(\Delta q), \Delta t; 0_{1 \times 3}, 1]$. Thus, the transformation of ego clip is $T^{clip} = \{ T_{init}, T_{init}T_{0}^{1}, ...,
T_{init}\prod_{i=1}^7 (T_{i}^{i+1})$. Then, we warp this toward 2D third view as $^{2D}t^{clip} = [T^{clip}_{0}[0:2, 3], ..., T^{clip}_{7}[0:2, 3]] - T^{clip}_{0}[0:2, 3]$. The loss used to regress the learning of the ego-downward transformation is,

\begin{equation}
    L(X^{t}) = || \c{B}^{trj} -  ^{2D}t^{clip}||_{L1},
\end{equation}

where $||||_{L1}$ denotes $L1$ norm as the loss.

\textbf{Third View Transformation Regression} $ \ $ The third view directly predicts the translation third view image, and the tracker bounding box center, $\c{B}^{trj}$ as output. It predicts frame-to-frame translation $\Delta ^{trd}t = (\Delta x, \Delta y)$. Thus, the output of a third view clip $Y^{flow} = \{ ^{trd}I^{flow}_{i}|i=1,...,7 \}$ is $^{trd-2D}t = [0, 0; \Delta x, \Delta y; ... ; \sum_{i=1}^{7}\Delta x, \sum_{i=1}^{7}\Delta y]$. We design the loss as,

\begin{equation}
  L(Y^{t}) = || \c{B}^{trj} -  ^{trd-2D}t||_{L1} .
\end{equation}

\textbf{ETVVT Model Learning} $ \ $  The four sub-channels intermediate layer features then concatenated into one feature vector as input for the discriminator which is a two-layered fully connected networks. The loss for verification regression is cross entropy loss to predict $true$ or $false$ and the sum of each sub-model losses,

\begin{equation}
    Loss = \sum_{i=0}^{1} y_{o,i} log(P_{o,i}) + L(X^{a}) + L(Y^{a}) + L(X^{t}) + L(Y^{t}),
\end{equation}

where $y$ is the binary indicator of prediction is correct and $p$ is the corresponding probability.

%-------------------------------------------------------------------------

\begin{figure*}[!th]
\begin{center}
   \includegraphics[width=1.0 \linewidth]{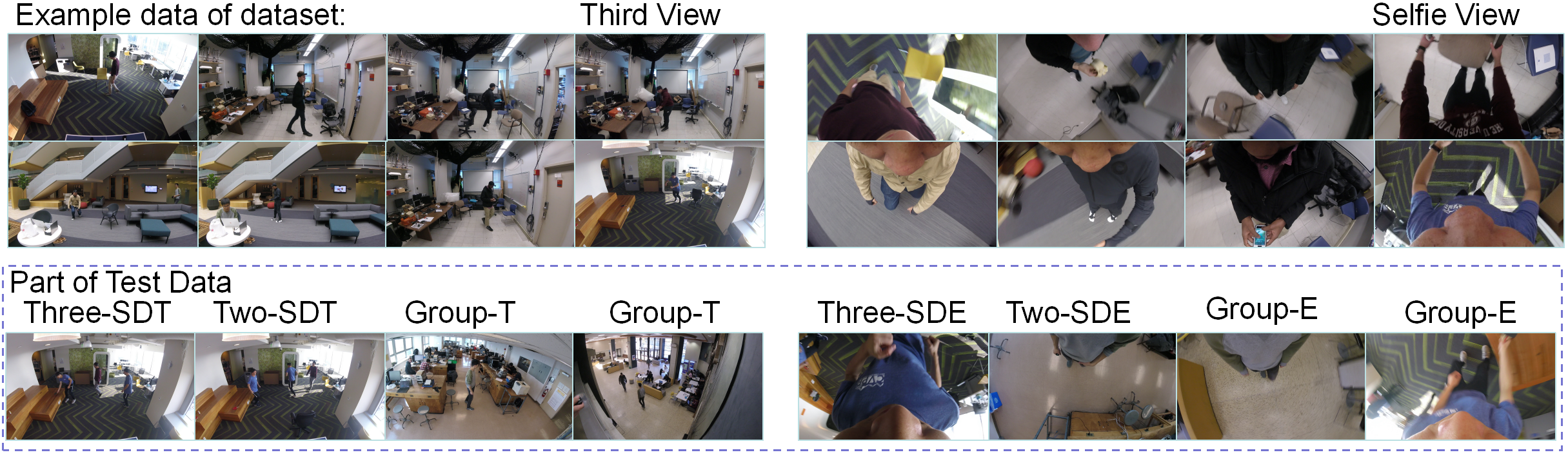}
\end{center}
   \caption{An illustration of part of our collected data. The training and validation data are collected under different backgrounds. For test data, we consider the same-dressing in third view (SDT) and also large area with over dozens of people. SDE denotes same dressing ego view.}
\label{fig:dataset_illustration}
\end{figure*}

\section{Experiments}
\label{Sec:experiments}
\subsection{Dataset Collection}

The dataset collection considers the following challenges: 1) same color dressing or close color; 2) background difference as context inference for verification; 3) number of people related with accuracy; 4) similar motion situation. All the data collected are listed in Table.\ref{table:Dataset_preparasion}, which contains a total number of $40$ videos. For the training and validation purpose, we collected $30$ single person ego-downward and third view videos under $5$ different backgrounds. For each pair, it contains an ego-downward video and a third view static video. For all the video pairs, we generate clips which contains $8$ raw images and $7$ flow images as training and testing purpose. We highlight the challenge of verification if the person in third view have the same dressing and collect extra data on this. The testing data contains $2$ to $3+$ person in view cases, and the synchronization is performed using GoPro camera remote controller.

%ToDo: Show part of the data

\begin{table}[t]
\caption{A summary of collected videos in our dataset.}
\begin{tabular}{ |p{2cm}|p{2.5cm}|p{3cm}| }
\hline
Single Person & Three backgrounds & A total $30$ pair of videos containing over $100,000$ image pairs\\
\hline
\hline
\multirow{5}*{Multi-person} & Two Person: No Crossing & $1$ pair of videos \\
\cline{2-3}
& Two Person: Crossing & $1$ pair of videos \\
\cline{2-3}
& Three Person: No Crossing & $1$ pair of videos \\
\cline{2-3}
& Three Person: Crossing & $1$ pair of videos\\
\cline{2-3}
& Group Crossing: & $4$ pair of videos \\
\hline
\hline
\multirow{2}*{Same Dressing} & Two Person: No Crossing & 1 pair of videos\\
\cline{2-3}
& Group $:$ Crossing & 1 pair of videos \\
\hline
\end{tabular}
\label{table:Dataset_preparasion}
\end{table}

\subsection{Implementation Details}
\label{Sec:datset}

\textbf{Dataset Preparation} For each pair of videos, we perform the following operations which can be repeated in a step by step manner: 1) parse the videos into images; 2) Generate dense optical flow and represent in $x$ and $y$ directional separate images \cite{opencvDenseFlow}; 3) For third view frames, first we perform person detection and tracking to obtain the bounding boxes \cite{Bewley2016_sort} for cropping. Then 3D pose estimation of generating the 3D joints is performed for each cropped image using HMR \cite{kanazawa2018end}; 4) The 3D poses set of each clip is then clustered using K-means algorithm \cite{kmeansclustering}, with $K=400$ in this paper. Then, we can obtain the action label of each frame. We also tried $300$, and $500$. It should be advised that a bigger $K$ should be more accurate for verification considering of a more general application purpose.

Following the above procedures, we can obtain: 1) raw image, flow images, and action label for ego-downward view; 2) raw image, flow images, bounding box, and action label of each person, and the corresponding 3D pose indicated by $19$ joints for third view (it is used to calculate the initial transformation $T = (R, t)$ for motion model). For all the $30$ single person videos, we choose $24$ for training and $6$ for testing.

%ToDo: Show PCA 3D plot of the Actions

\textbf{Training Details} We choose to initialize each model using a pre-trained ResNet \cite{he2016deep} which is trained on ImageNet-ILSVRC \cite{russakovsky2015imagenet}. All the models are implemented in Pytorch \cite{paszke2017automatic}, with learning rate as $0.01$ and weight decay $0.001$ for $200$ epochs using two Nvidia 1080 GPUS. For our network, we trained each sub-model independently. Then we perform joint optimization for final verification.

\begin{figure}[!th]
\begin{center}
   \includegraphics[width=1.0\linewidth]{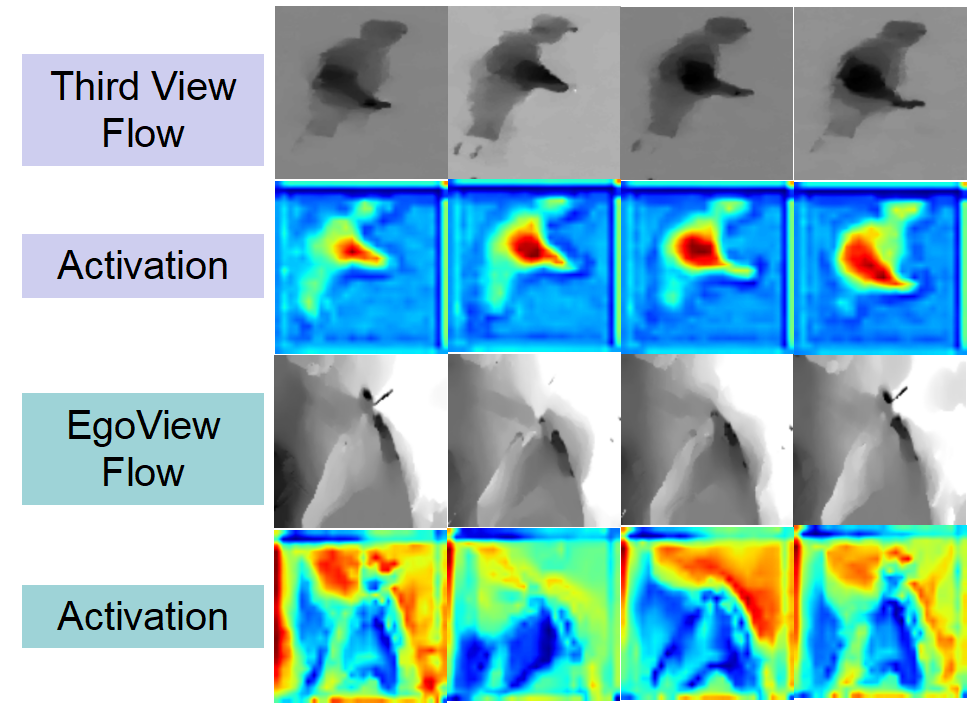}
\end{center}
   \caption{The motion model Block 3 activations. The colors range from blue to red, denoting low to high activations.}
\label{fig:translation_activation}
\end{figure}

\subsection{Results and Comparison}

\textbf{Baselines} $ \ $ We first implement multiple baselines to compare the performance considering inputs, and models. These baseline method are proposed in peer researches \cite{fan2017identifying,sigurdsson2018actor, qiu2017learning} including spatial-domain siamese network \cite{fan2017identifying}, motion-domain siamese network\cite{fan2017identifying}, two-stream semi-siamese network \cite{fan2017identifying}, triplet network \cite{sigurdsson2018actor}, and temporal domain image and flow network \cite{fan2017identifying, qiu2017learning}. We also demonstrate the weight share performance for siamese-network. We deploy 2D and 3D Resnets \cite{qiu2017learning} to learning spatial and temporal features.

\begin{table}[t]
\caption{Verification accuracy (in $\%$) baselines on our dataset, and higher is better. Where SW denotes share weight.}
\begin{tabular}{ |p{2.8cm}|p{1cm}|p{1cm}| p{1cm}|p{1cm}|}
\hline
& Resnet-18 &  Resnet-34 & Resnet-50 & Resnet-101 \\
\hline
Siamese Image& 50.39 &	51.03&	50.55 &	50.42 \\
\cline{1-1}
Siamese Flow & 52.53 &	50.75&	51.63 &	\textbf{52.06} \\
\cline{1-1}
Semi-siamese SW & 53.34 &	52.41 & 52.78 &	51.35 \\
\cline{1-1}
Semi-siamese & 52.1 &	51.89 &	51.29 &	50.91 \\
\hline
\hline
Temporal-Siamese Image  & 52.21  &	51.6  &	51.43 & - \\
\cline{1-1}
Temporal-Siamese Flow & \textbf{54.77} & \textbf{55.9} &	\textbf{55.10} & - \\
\cline{1-1}
Temporal Semi-siamese&  51.74& 	53.96& 	50.89& - \\
\hline
\hline
Triplet \cite{sigurdsson2018actor}&  52.80& 51.28 & 51.63	& 51.49 \\
\hline
\end{tabular}
\label{table:baselines_of_other_methods}
\end{table}

For feature consideration, we performed the training and testing using image data and flow data in independent network, while we also performed learning using both information in a semi-siamese approach. Table.\ref{table:baselines_of_other_methods} summaries the accuracies of the above models. In this paper, we use accuracy as metric to evaluate the models as \cite{sigurdsson2018actor}. It shows in the table that temporal models are significantly much better for our tracking problem, and also flow information is more accurate. It is due to our dataset requires person to move frequently and fast, thus makes it hard to verify using pure context feature. The the maximum accuracy according to these methods is $55.9 \%$ which is 3D temporal Resnet-34 model using optical flow as input. However, the semi-temporal model does not show any improvement, which may caused by limited data of color feature of our dataset.

In this table, we can also see that a share weight siamese-model is more effective then the none-share models with an average $1\%$ percent higher. For Semi-siamese model, in spatial domain, it is a four channel network takes both flow and image as input. The triplet model is implemented as proposed in paper \cite{sigurdsson2018actor}, where a none-corresponding image is used input of the model. The result accuracy indicates that the triplet structure can achieve similar performance compared to temporal flow model, and it does not require huge amount of parameter to train.

For the base line implementation, we did not implement semi-triplet as proposed in \cite{fan2017identifying} since we regard the tracking is performed in large crowd. Thus, the semi-triplet model will have to perform exponential times of verification due to the requirement of input. However, the above data tells the following learning: 1) flow information is more important for localization; 2) complex model may not help if simply use spatial and temporal information.

\textbf{ETVVT Model Testing}

\begin{figure*}[!th]
\begin{center}
   \includegraphics[width=1.0\linewidth]{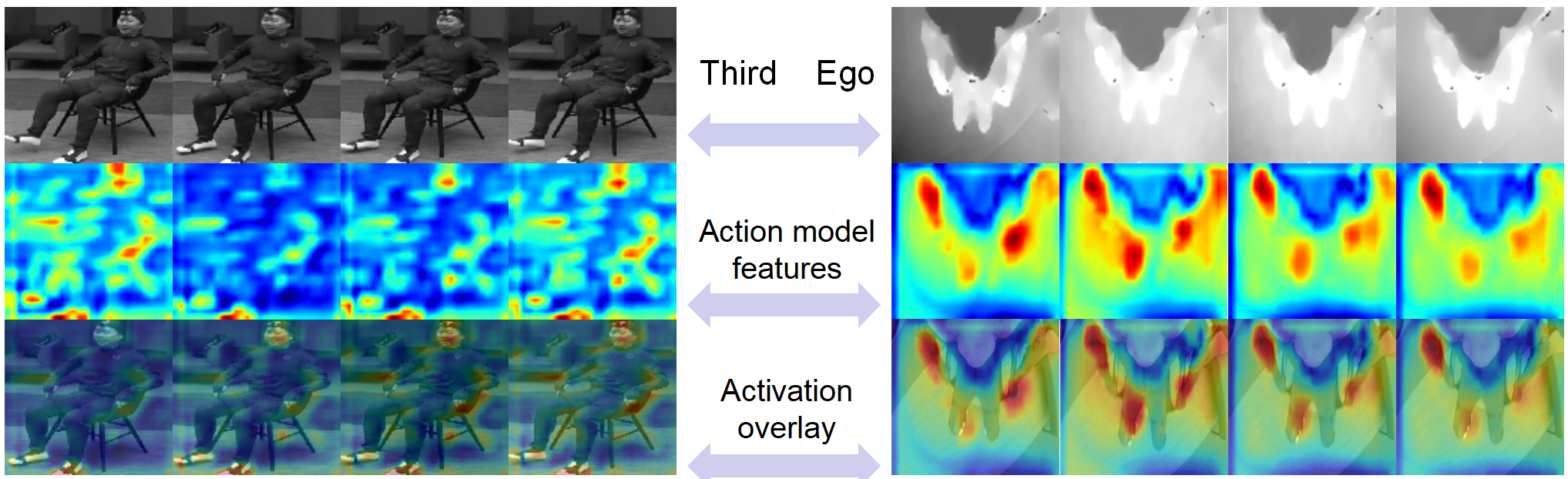}
\end{center}
   \caption{Block 3 activations of action model. The colors range from blue to red, denoting low to high activations.}
\label{fig:action_activation}
\end{figure*}

\textbf{1) Performance and Analysis} $  \ $ We also test our proposed model on the single person dataset. The results are summarized in Table.\ref{table:proposed_medtho_accuracy}, where we also test the action model and motion model separately. We can obtain that the proposed method output performs the best base line by $18.32 \%$. The independent action model can achieve $72.5 \%$ in accuracy and translation model can achieve $70.03 \%$ in accuracy.

\begin{table}[t]
\caption{Verification performance of proposed model. AP($\%$): Average Precision, and AR$\%$: Average Recall}
\begin{tabular}{ |p{2.8cm}|p{1.6cm}|p{1cm}|p{1cm}|}
\hline
Model & Accurcy $\%$ & AP & AR   \\
\hline
Action Model   &  72.5& 68.92 &  42.32\\
Translation Model & 70.03 & 64.38& 38.74  \\
ETVVT Model & 74.22  & 69.78 & 47.93\\
\hline
\end{tabular}
\label{table:proposed_medtho_accuracy}
\end{table}

%ToDo:do we need loss??

\textbf{2) Action VS Motion Model} $ \ $ The result shows that Action model has a $2.47 \%$ higher accuracy than Motion model, and $4.54 \%$ higher average precision. It is because the motion model does not tell any difference when human is static or just move the part of the body. We also visualize the activations and the overlay to image of motion model as illustrated in Fig.\ref{fig:translation_activation}. It can be seem that the third view translation highly attend to the center of the flow, while, the ego motion model attend to the outer body region for translation estimation. For action sub-model, the activations of each model the third block is Fig.\ref{fig:action_activation}. We observe the action model attending to joints to perceive pose information both in RGB-image and flow images.

\textbf{3) Ego Odometry VS Third View Odometry} $ \ $  We also compare the importance of ego-view translation and third view translation. We directly introduce to add the translation as an independent channel into the temporal semi-siamese model, in a fully connected layer (Appear In appendix). The result shows that third view translation can increase the validation accuracy ($20 \%$ of the training data) from $79.05 \%$ to $81.80 \%$. It can be explained according to Fig.\ref{fig:translation_activation} that our ego view has limited view of world, also the head motion introduces error.

\begin{table}[t]
\caption{The verification accuracy $\%$ on multi-people testing data.}
\begin{tabular}{ |p{1.9cm}|p{2.5cm}|p{1.3cm}|p{1.2cm}| }
\hline
& Test Case & Accuracy & Bayes Filter \\
\hline
\multirow{5}*{Multi-person} & Two Person $:$ No Crossing &  72.26 & 96.17 \\
\cline{2-4}
& Two Person $:$ Crossing & 62.18 &  80.76\\
\cline{2-4}
& Three Person $:$ No Crossing & 72.25 & 92.27 \\
\cline{2-4}
& Three Person $:$ Crossing & 65.39 & 91.52\\
\cline{2-4}
& Group Crossing $:$ & 57.26 & - \\
\hline
\hline
\multirow{2}*{Same Dressing} & Two Person $:$ No Crossing & 72.26 & 96.17\\
\cline{2-4}
& Three Person $:$ Crossing & 65.39 & 91.52\\
\hline
\end{tabular}
\label{table:testing_with_multiple_people_data}
\end{table}

\textbf{4) Test On Multi-person Videos} Then, we test the proposed model in our multi-moving people cases with results illustrate in Table.\ref{table:testing_with_multiple_people_data}. For the ground truth, we use the the tracker and human label to obtain. It is can be seem in Table.\ref{table:testing_with_multiple_people_data} that ETVVT can achieve an average accuracy $67.767 \%$ for all the test cases. For group cross, the filtering fails since to much crossing happens.For implementation, we perform prediction of all the detected person and conclude based on the maximum score.

ETVVT model has lower accuracy when the ego-camera mounted person crossed with other pedestrian. It is due to partial observable of the body, the 3D pose estimation would fail. In this paper, we also introduce a Bayes filter with velocity prediction to filter the verification results\cite{ababsa2011robust}. The filterred result are illustrated in Table.\ref{table:testing_with_multiple_people_data}, which shows promising in few person in view scenario.

\textbf{ETVVT Adaptivity Analysis}

\begin{figure}[!th]
\begin{center}
   \includegraphics[width=0.9\linewidth]{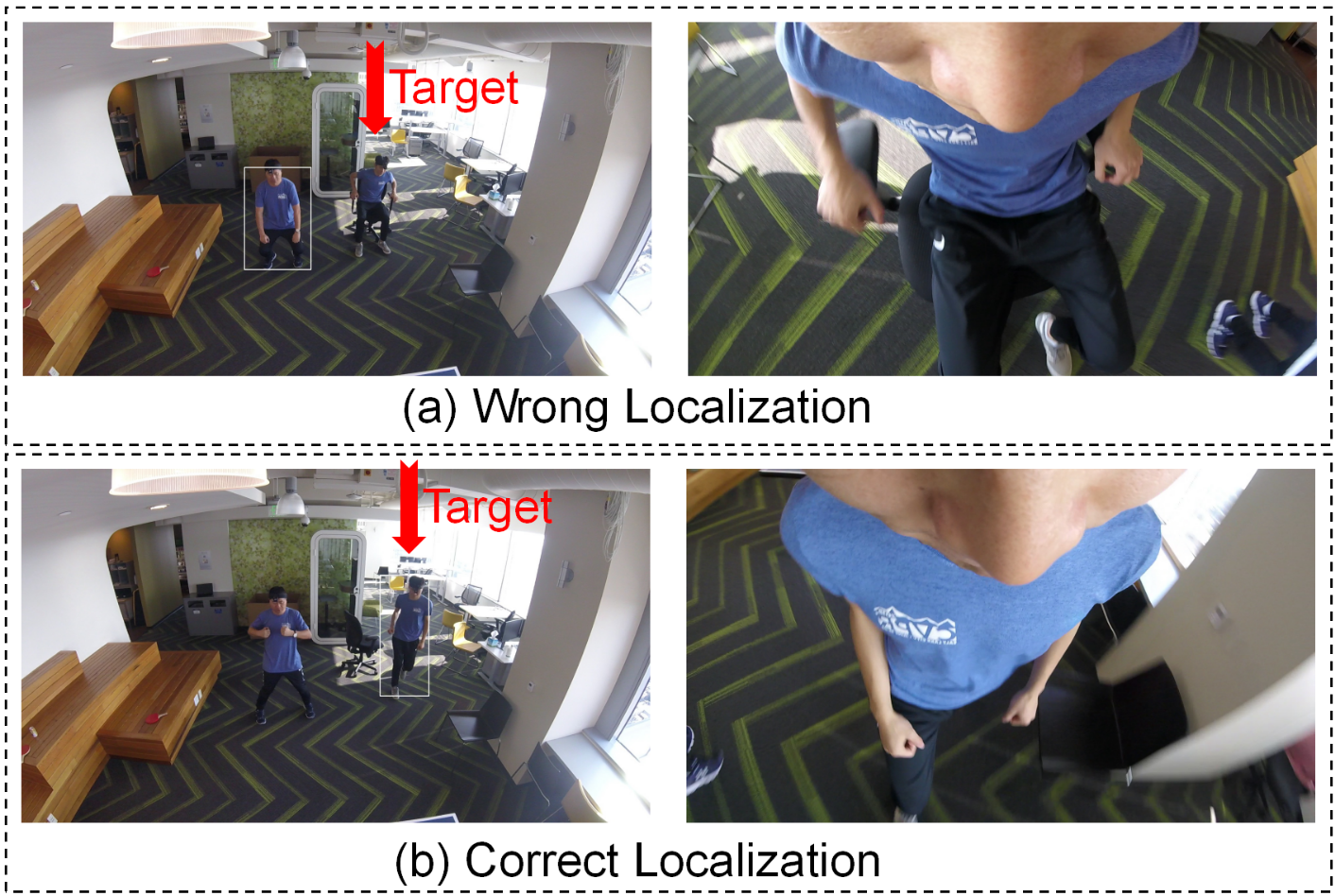}
\end{center}
   \caption{Left side are third view frames and right are ego view frames. The white rectangle denotes localized result. The target perform with ego camera has a red arrow on head.}
\label{fig:analysis_wrong_correct}
\end{figure}

Our model directly transforms ego view information into the third view coordinate system, and we firstly introduced 3D pose to perform understanding. The geometry and action information model help to learn the two view pose and motion information for cross view verification. Besides, we use short-term video clip as input which enables on-line processing.

We also find several limitation of our model at current stage. First, if all the person are static or with similar pose in view, our algorithm would fail. Second, if all person with the same action and motion, it also fails (in Fig.\ref{fig:analysis_wrong_correct}). It is illustrated in Fig.\ref{fig:analysis_wrong_correct}(a), the two person have the same dressing and doing the same motion, it localized the wrong person in view. However, in most time, the person are with different motion and action (in Fig.\ref{fig:analysis_wrong_correct}(b)), our model can obtain the correct result.

\section{Conclusion}
We present an action and motion learning model for cross view localization and tracking via introducing 3D pose as transformation for alignment. It is motivated by observation that the ego view is not able sense the third view absolute coordinate information. Our experimental results show that our method outperforms the state-of-art verification model on cross view verification, even with same dressing. It delivers a competitive generalization of cross view verification on semi-supervise learning for localization and tracking using action and motion clue.

{\small
\bibliographystyle{ieee}
\bibliography{egbib}
}

\section{Ego Odometry VS Third View Odometry}

In this Section, we provide the network architecture for supplementary of comparison described in Section 4.3.
\begin{figure}[!th]
\begin{center}
   \includegraphics[width=1.0 \linewidth]{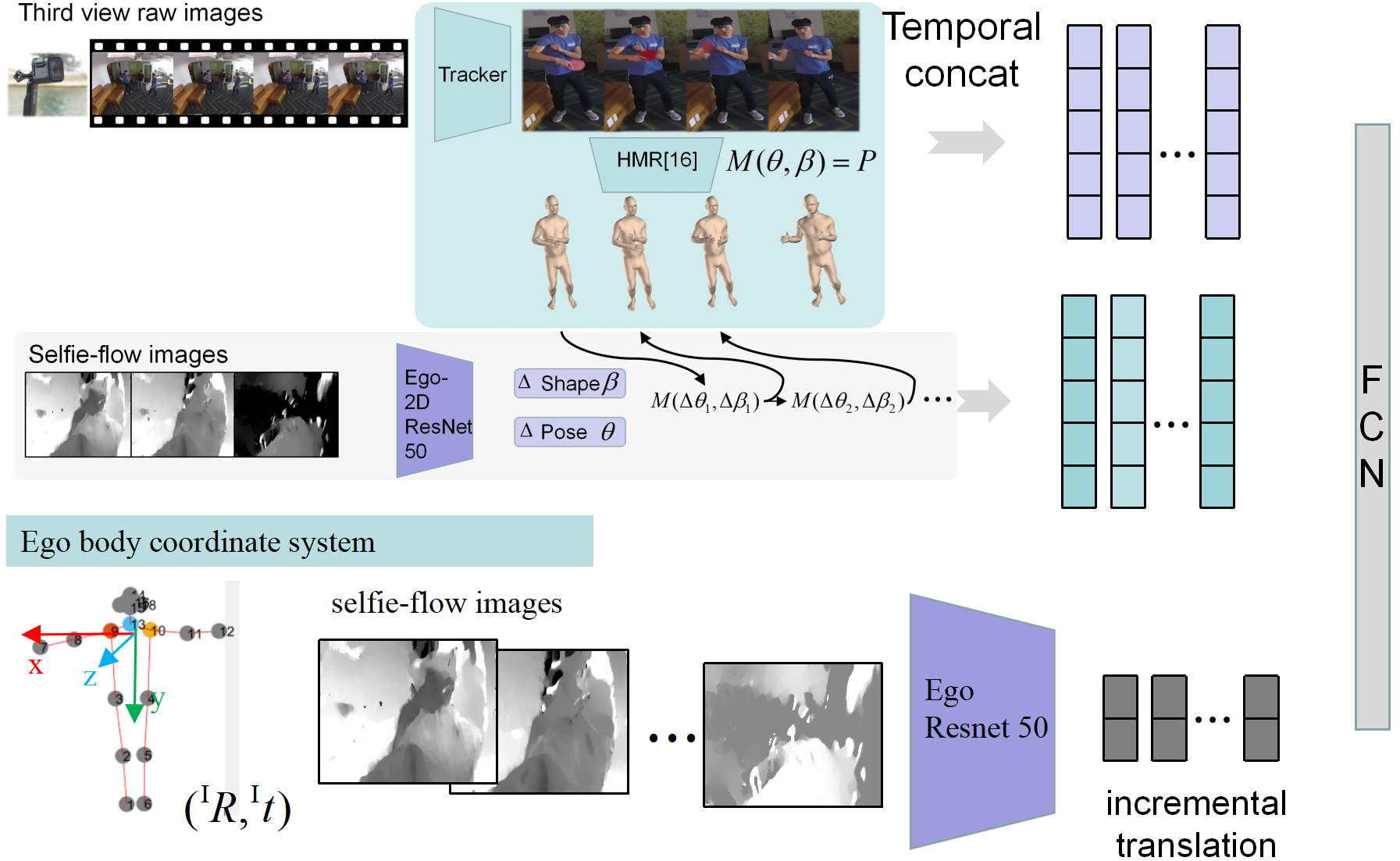}
\end{center}
   \caption{The ego-translation only with action semi-siamese network for cross view validation.}
\label{fig:egoView_translation_model}
\end{figure}

\textbf{Ego View Odometry Network Architecture}

It has been discussed in Section 3.2, the model of ego-view translation has been discussed. A ego view translation model needs input: 1) the initial transformation in third view as described in Section 3, that is, $T_{init}=(^{T}R, ^{T}t)$. 2) the consecutive flow frames. The output is $^{trd}t=[0, 0; \Delta x, \Delta y; ...;\sum_{i=1}^{7}\Delta x, \sum_{i=1}^{7}\Delta y]$ which is already transformed into third view coordinate system and concatenated together.

To compare the importance of the odometry information, we further introduce the ego-translation only model which is illustrated in Fig.\ref{fig:egoView_translation_model}.

\begin{figure}[!th]
\begin{center}
   \includegraphics[width=1.0 \linewidth]{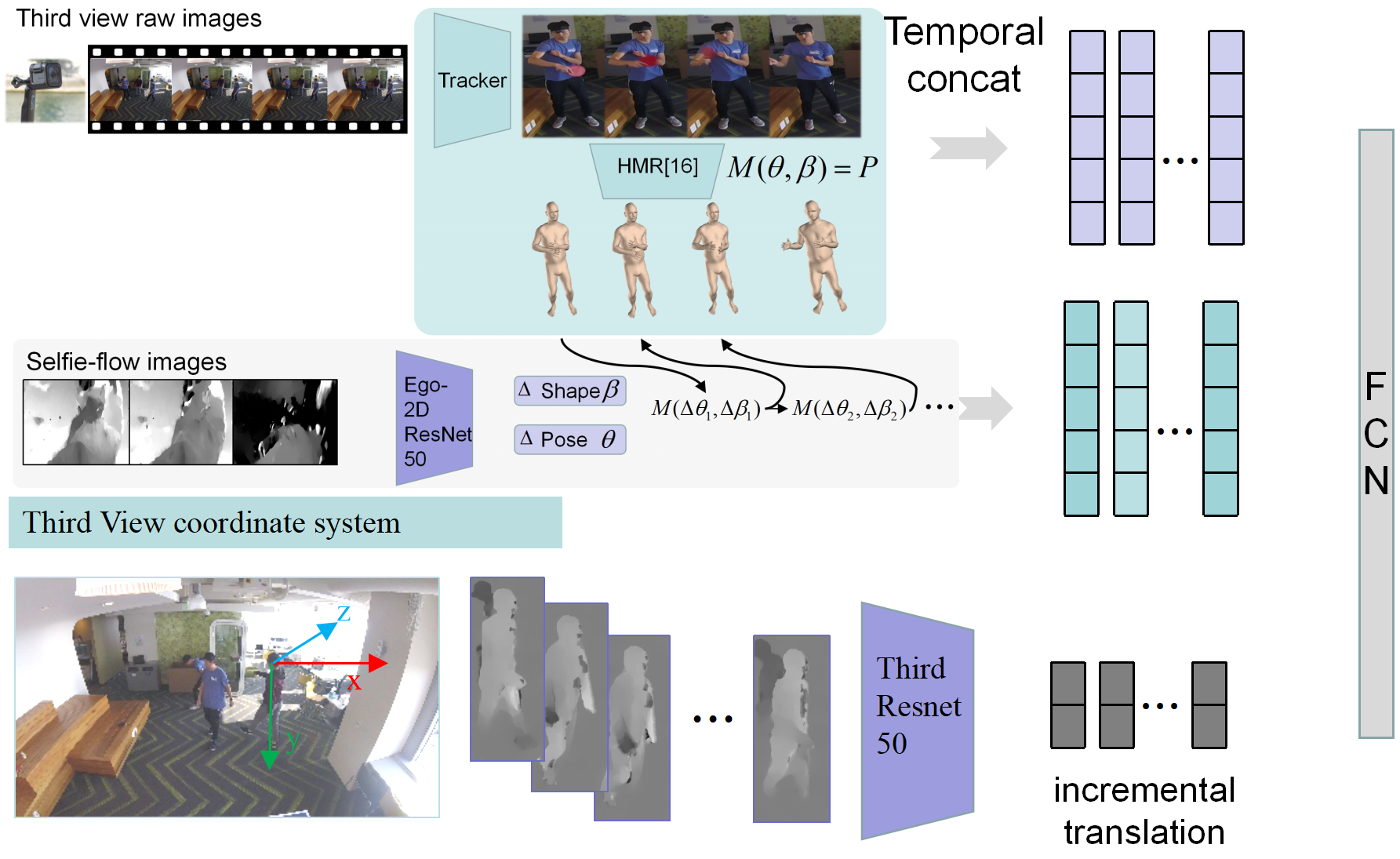}
\end{center}
   \caption{The third-view translation only with action semi-siamese network for cross view validation.}
\label{fig:thirdView_translation_model}
\end{figure}

\textbf{Third View Odometry Network Architecture}
The third view odometry information only model is illustrated in Fig.\ref{fig:thirdView_translation_model}. For the third view translation prediction model, the input is $7$ consecutive flow frames. It directly outputs $^{trd-2D}t = [0, 0; \Delta x, \Delta y; ... ; \sum_{i=1}^{7}\Delta x, \sum_{i=1}^{7}\Delta y]$, which should indicate the translation of person center in third view image.

\section{Prepare Training and Testing Dataset}
We also provide the detailed procedures to generate dataset. The general procedures are:
\begin{enumerate}
\item Parse the videos into images
\item Generate dense optical flow and represent in $x$ and $y$ directional separate images \cite{opencvDenseFlow}
\item For third view frames, first we perform person detection and tracking to obtain the bounding boxes \cite{Bewley2016_sort} for cropping. Then 3D pose estimation of generating the 3D joints is performed for each cropped image using HMR \cite{kanazawa2018end}
\end{enumerate}

\subsection{Action Models}
Action models has to train the model to recognize the action from both view and then perform verification. The dataset is prepared as follows:

\begin{enumerate}
\item We concatenate every consecutive $8$ (time $t_{0},...,t_{7}$) 3D poses in a vector. Then, we perform K-means to do clustering \cite{kmeansclustering}, with $K=400$ in this paper. The K-means index is the action label of the last frame of each $8$ frames.
\item From all the labeled images, we randomly select $80 \%$ for training and $20 \%$ for testing.
\item For ego view data, we bundle the initial corresponding third view image (time $t_{0}$) and the consecutive $7$ ego view flow images (time $t_{1},...,t_{7}$). For third view data, we bundle the consecutive $8$ RGB-images (time $t_{0},...,t_{7}$).
\end{enumerate}

\subsection{Odometry Models}

Odometry models use the third view bounding box center translation as output, that is, $\c{B}^{trj} = \{\c{B} - \c{B}_{0} \}$ as described in Section 3.2. At the initial independent training stage, we follows:

\begin{enumerate}
\item Calculate the third view person translation $\c{B}^{trj}(t_{k}) = \{ \c{B}(t_{k}-t_{k-i})- \c{B}(t_{k}-t_{k-7}) | i=7, 6, .., 0\}$ at time $t_{k}$
\item For ego view, we first calculate the initial pose,  $T_{init}=(^{T}R, ^{T}t)$, according to Section 3.2. Then, we bundle the transformation $T_{init}$ and $7$ ego view flow images at time $t_{k-6},...,t_{k}$.
\item For third view, we bundle the $7$ consecutive flow images at time $t_{k-6},...,t_{k}$ as training input.
\end{enumerate}

\section{Video}

We provide a video to demonstrate of ETVVT performance. We show the following cases:

\begin{enumerate}
\item The model verification for localization and tracking of three person in view, with two person are with the same dressing and mounting.
\item The comparison of using filter and raw model prediction
\item The three person in view and crossing case with two person are with the same dressing and mounting.
\item Large group case with: 1) only one person is moving; 2) several person are moving.

\end{enumerate}

\end{document}